\documentclass[10pt]{article} 
\usepackage[preprint]{other/tmlr}

\usepackage{amsmath,amsfonts,bm}

\def\eqref#1{equation~\ref{#1}}

\def\1{\bm{1}}

\DeclareMathAlphabet{\mathsfit}{\encodingdefault}{\sfdefault}{m}{sl}
\SetMathAlphabet{\mathsfit}{bold}{\encodingdefault}{\sfdefault}{bx}{n}

\usepackage{hyperref}
\usepackage{url}
\usepackage{graphicx}
\usepackage{cleveref}
\usepackage{booktabs}
\usepackage{subcaption}
\usepackage{soul}
\soulregister\cite7
\usepackage{xcolor}
\sethlcolor{yellow}
\usepackage{changes}

\usepackage{AbdAlmageed_style}

\crefname{figure}{Fig.}{Figs.}
\crefname{section}{Section}{Sections}

\title{Causal Representation Learning on High-Dimensional Data: \\Benchmarks, Reproducibility, and Evaluation Metrics}

\author{\name Alireza Sadeghi \email alirezs@clemson.edu \\
      \addr Holcombe Department of Electrical and Computer Engineering\\
      Clemson University
      \AND
      \name Wael Abd Almageed \email wabdalm@clemson.edu \\
      \addr Holcombe Department of Electrical and Computer Engineering\\
      Clemson University
      }

\def\month{MM}  
\def\year{YYYY} 
\def\openreview{\url{https://openreview.net/forum?id=XXXX}} 

\begin{document}

\maketitle

\begin{abstract}

Causal representation learning (CRL) models aim to transform high-dimensional data into a latent space, enabling interventions to generate counterfactual samples or modify existing data based on the causal relationships among latent variables. To facilitate the development and evaluation of these models, a variety of synthetic and real-world datasets have been proposed, each with distinct advantages and limitations. For practical applications, CRL models must perform robustly across multiple evaluation directions, including reconstruction, disentanglement, causal discovery, and counterfactual reasoning, using appropriate metrics for each direction. However, this multi-directional evaluation can complicate model comparison, as a model may excel in some direction while under-performing in others. Another significant challenge in this field is reproducibility: the source code corresponding to published results must be publicly available, and repeated runs should yield performance consistent with the original reports. In this study, we critically analyzed the synthetic and real-world datasets currently employed in the literature, highlighting their limitations and proposing a set of essential characteristics for suitable datasets in CRL model development. We also introduce a single aggregate metric that consolidates performance across all evaluation directions, providing a comprehensive score for each model. Finally, we reviewed existing implementations from the literature and assessed them in terms of reproducibility, identifying gaps and best practices in the field.

\end{abstract}

\section{Introduction}
Deep neural networks generally try to discover the correlations between input and output data, and make inference based on the learned correlations \cite{pearl2019seven, jiao2024causal, huang2025visual, kumar2023causal, wu2023discover, izmailov2022feature}. However, it is probable that these models latch to spurious correlations, which in turn result in overfitting \cite{zhang2015multi}, generalization \cite{truong2022generalization}, or suboptimal performance in data generation models \cite{choi2020fair}.
To this end, a direction of studies have formed to develop deep neural networks to first discover the underlying cause-effect relationships and then utilize them, instead of statistical correlations, in various application domains 
\cite{tesch2023causal, castro2020causality, sharma2024incorporating} including disentangling models.

Disentanglement of latent variables, roughly defined as assigning each generative factor to a distinct latent variable, has received significant attention in recent years. Achieving effective disentanglement facilitates more controllable generation of new samples or modifying the current ones by manipulating the corresponding latent variables \cite{pastrana2022disentangling}. Previous research \cite{khemakhem2020variational,locatello2019challenging} has shown that disentangling latent variables using only unsupervised methods is fundamentally impossible due to identifiability constraints, which makes these techniques inherently unreliable for this task. Consequently, to uniquely identify causal variables and their underlying structure, it is essential to incorporate prior knowledge and/or assumptions about the data-generating process or apply supervision \cite{eberhardt2016green, locatello2019challenging}. 

Initial research efforts for disentanglement \cite{higgins2017beta,burgess2018understanding,kumar2017variational} assume the latent factors are mutually independent, attempting disentanglement based on this premise. However, real-world variables typically exhibit causal relationships rather than independence. It has been demonstrated that approaches relying on independence priors struggle to disentangle causally interconnected factors \cite{shen2022weakly}. Consequently, causal representation learning (CRL), which aims at uncovering high-level causal variables from low-level observational data, has become a prominent research focus \cite{scholkopf2021toward,rajendran2024causal, lv2022causality, wang2022causal, ahuja2023interventional}. Causal disentanglement is particularly valued for its invariance, enhancing robustness against distribution shifts \cite{scholkopf2022causality, arjovsky2019invariant}. Furthermore, it enables causally-controllable generation, allowing data production under various desired interventions on latent factors. Central to this causal inference approach are Directed Acyclic Graphs (DAGs), which formally represent causal relationships graphically. DAGs facilitate intervention techniques such as the back-door and front-door criteria, assisting researchers in adjusting for confounding variables and accurately estimating causal effects \cite{jiao2024causal}. DAGs can also be helpful for counterfactual reasoning, the third rung of the ladder of the causation \cite{pearl2018book}, as it is possible to track the impact of intervention applied to one variable on other variables. Beyond identifying interventional effects, structural causal models grounded in DAGs also support counterfactual reasoning, the third rung of the ladder of the causation \cite{pearl2018book}, as it extends interventions by conditioning on observed outcomes and asking how those outcomes would have differed under alternative interventions. Counterfactual reasoning using deep neural networks is critical in different fields where conducting an actual experiment is challenging maybe due to feasibility or ethical problems.

Similar to the development of other deep learning models, CRL models must be trained on appropriately designed datasets. However, unlike conventional supervised deep learning models that rely on data samples paired with target labels, CRL models require explicit information about the underlying cause–effect relationships among variables, as well as the corresponding variable labels or values for each data sample. To ensure robustness and generalizability, such datasets should exhibit well-defined structural properties and a certain degree of complexity. Although various synthetic and real-world datasets have been proposed and employed in existing studies, many of them lack several essential characteristics necessary for comprehensive and effective causal representation learning.

Once a CRL model is developed, it should be evaluated from multiple perspectives (\eg reconstruction ability, counterfactual reasoning, etc.); not only to verify that its performance meets acceptable standards, but also to enable fair comparison across different CRL frameworks. While several evaluation metrics have been introduced to assess different aspects of CRL performance, to the best of our knowledge, there is currently no unified metric capable of integrating and summarizing all these criteria into a single quantitative score for holistic model comparison.

In this work, we conduct a detailed review of the existing literature in the field of CRL model development. First, we discuss the desirable characteristics of an ideal dataset for CRL model development and use these criteria to critically evaluate the existing synthetic and real-world datasets currently utilized in this domain. We then examine the key evaluation dimensions that should be considered when assessing a CRL model, along with the corresponding performance metrics applied in each dimension. Recognizing that most existing studies report evaluations on only a subset of these essential  dimensions, and that even when multiple metrics are reported they often yield conflicting assessments, we propose an integrated evaluation framework that aggregates multiple metrics into a single, unified score. This unified metric is designed to optimally combine the performance indicators across different evaluation dimensions, thereby providing a comprehensive quantitative measure of a CRL model’s overall effectiveness relative to other models. While this unification approach have been applied in other research domains, this study represents the first attempt to introduce such a unified scoring methodology for CRL models.
\section{Synthetic and Real World Datasets}

\subsection{Expected Features in CRL Datasets} 
\label{sub: expected_features}

The real world encompasses a multitude of causal factors, confounders, and inherent uncertainties. Therefore, an ideal CRL model—expected to operate effectively in such complex environments—should be trained on datasets that closely reflect these real-world characteristics. Consequently, the datasets used for CRL model development should incorporate features that capture the structural and statistical properties of real-world causal systems as faithfully as possible.
Although it is challenging to exhaustively identify all potential characteristics observed in real-world scenarios, we propose several fundamental properties that are both evident and essential to consider when designing new real-world datasets or generating synthetic ones. These characteristics include: (i) inclusion of various basic causal junctions (ii) a sufficient number of variables, (iii) alignment with the real-world semantics , and (iv) the presence of confounders.

\begin{figure*}
    \centering
    \includegraphics[width=0.5\linewidth]{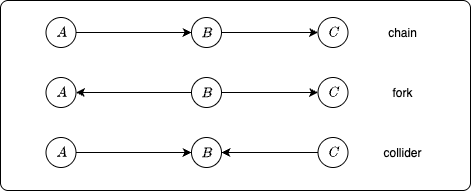}
    \caption{Fundamental causal
junction types that constitute the building blocks of causal graphs.}
    \label{fig:basic_junctions}
\end{figure*}

\textbf{Basic causal junctions:} 
There are three basic causal junction structures \cite{pearl2018book}—chain, fork, and collider—as illustrated in \cref{fig:basic_junctions}. Each of these junction types exhibits distinct statistical properties. In a chain structure ($A \to B \to C$), variables $A$ and $C$ are statistically correlated, but conditioning on the intermediary variable $B$ renders them conditionally independent. Similarly, in a fork structure ($A \gets B \to C$), variables $A$ and $C$ are correlated due to their common cause $B$, and conditioning on $B$ removes this dependence, making them conditionally independent. Conversely, in a collider structure ($A \to B \gets C$), variables $A$ and $C$ are initially independent, but conditioning on the collider variable $B$ induces a dependence between them. These three junction types are fundamental building blocks of real-world causal systems. Therefore, a robust CRL model should be capable of correctly identifying and reasoning over each of them. To this end, datasets used for training CRL models should explicitly include examples representing all three causal junction types to ensure comprehensive learning of causal dependencies.

\textbf{Sufficient number of variables:} In real-world scenarios, a large number of observable causal variables $X=\{X_1, X_2, ..., X_n \} $ interact alongside various sources of noise $\epsilon$ and latent confounders $U$, making it challenging for a CRL model to perform effectively in practical applications. However, if a CRL model is trained on datasets that include an adequate and diverse set of variables, it can better learn to analyze complex dependencies and adapt to real-world conditions, ultimately achieving higher generalization and performance. Therefore, an ideal dataset should incorporate multiple variables, as increasing the number of variables typically enables the CRL model to capture more complex causal relationships during practical deployment. Nevertheless, determining the optimal number of variables involves balancing model reliability and computational feasibility. While adding more variables generally improves the model’s capacity to represent intricate causal mechanisms, it also increases the dataset’s complexity, leading to longer training times and higher computational costs, which may make deployment impractical. Consequently, a trade-off must be established between the dataset’s dimensionality and the available computational resources, depending on the specific objectives of the study. Additionally, the feasibility of collecting data for certain variables is an important consideration. In many cases, it may not be possible to directly measure all relevant causal factors. However, if all variables can be observed and the only limitation concerns training time and computational cost, we suggest researchers prioritize retaining variables that is causally proximal to the objective of the study. If dimensionality reduction is required, variables with indirect effects—whose influence is mediated through other directly related variables—can be considered for elimination. For instance, in a chain $X_i\to X_j \to Y$ (where $Y$ is the objective of the study), the variable $X_i$ exerts an indirect effect on $Y$ that is mediated by $X_j$. Conditioning on the proximal variable $X_j$ renders $X_i$ and $Y$ conditionally independent, suggesting that $X_i$ may be pruned to reduce complexity without losing the immediate generative mechanism of $Y$. Transitioning from a complete causal graph $\mathcal{G}$ to a simplified subgraph $\acute{\mathcal{G}}$ requires a trade-off: while $n$ is reduced to optimize training throughput, the set must remain sufficient to avoid introducing omitted variable bias. Thus, researchers should prioritize variables with direct structural paths to the objective to maintain model reliability under resource constraints.

\textbf{Real-world semantic alignment:} The dataset used for training a CRL model should closely reflect real-world scenarios to enable the model to generalize effectively and perform reliably in practical applications. This characteristic is particularly critical when generating synthetic datasets. Synthetic data should emulate real-world phenomena rather than being purely hypothetical or artificial. Therefore, constructing synthetic datasets based on physically meaningful variables, realistic concepts, and equations derived from real-world systems is far more valuable than producing data that lacks physical grounding. Datasets synthesized with attention to the underlying physical or causal principles enable CRL models to develop more robust and generalizable representations, improving their performance when deployed in real-world environments that are similar—but not identical—to the training setting. Furthermore, an ideal dataset, whether real-world or synthetic, should include both numerical and categorical variables, as both types naturally occur in real systems. Consequently, a well-designed CRL model must be capable of effectively handling and reasoning over these heterogeneous data types.

\textbf{Confounders:}
Confounders—usually unobservable or unmeasured variables that simultaneously influence both the cause and the effect—are pervasive and often unavoidable in real-world systems. These hidden factors can distort causal inference by introducing spurious associations between variables. Therefore, an ideal CRL model should be capable of identifying and accounting for the presence of confounders to ensure accurate causal representation. To facilitate this capability, the datasets used for training should explicitly incorporate confounding effects, either through observed variables that play confounding roles or by simulating unseen confounders. In this way, the model can learn to disentangle true causal relationships from confounded correlations, enhancing its robustness and reliability in real-world applications.

With these characteristics established, we next turn to an examination of the real-world and synthetic datasets currently employed in CRL research. Specifically, we assess how well these datasets align with each of the aforementioned criteria, analyzing their capacity to reflect the required causal structure, semantic consistency, and evaluation coverage. This analysis allows us to identify systematic strengths and limitations in existing benchmarks, and to clarify the extent to which current datasets support meaningful and comprehensive evaluation of CRL models.

\begin{figure}
    \begin{subfigure}[c]{0.45\textwidth}
    \centering
    \includegraphics[width=\textwidth]{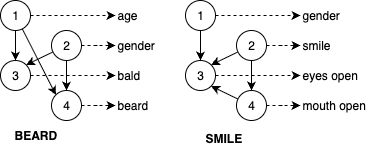}
    \caption{Causal graph for CelebA(BEARD) and CelebA(SMILE) datasets}
    \label{fig:CelebA}
    \end{subfigure}
    \hfill
    \begin{subfigure}[c]{0.45\textwidth}
    \centering\includegraphics[width=0.7\textwidth]{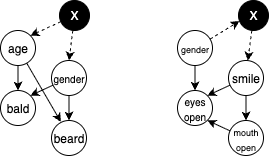}
    \caption{Possible confounders for CelebA(SMILE) and CelebA(BEARD) datasets}
    \label{fig:poss_conf}
    \end{subfigure}
    
    \vspace{1cm}
    \begin{subfigure}[c]{\textwidth}
    \centering
\includegraphics[width=\textwidth]{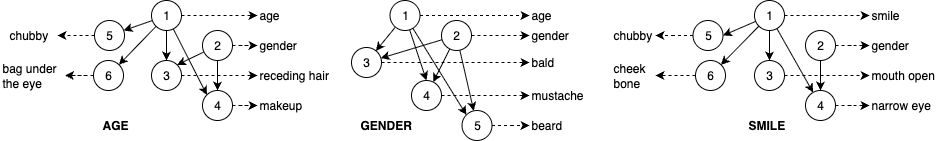}
    \caption{Three other subsets of CelebA dataset for CRL development}
    \label{fig:celebA2}
    \end{subfigure}
    \caption{a) Causal graphs for CelebA(SMILE) and CelebA(BEARD), two commonly used subsets of the CelebA dataset for CRL model development. b) Illustration of an unseen variable $X$ acting as a potential confounder, which may explain the dependencies observed between \textit{age} and \textit{gender} in CelebA(BEARD), and between \textit{gender} and \textit{smile} in CelebA(SMILE). c) Causal graphs for additional CelebA subsets constructed and utilized by \cite{huang2025visual}, with subset names consistent with those used in their original study.}
\end{figure}

\subsection{Real World Datasets}
\label{sub: real-world}
Although MorphoMNIST was used as a real-world dataset for CRL development in \cite{komanduri2024causal}, featuring a simple causal graph ($z_1 \to z_2$) where $z_1$ represents digit thickness and $z_2$ represents intensity, the majority of studies in this field have relied on the CelebA dataset for real-world CRL experiments. CelebA \cite{liu2015deep} comprises face images of 10,000 individuals, with roughly 20 images per individual, totaling more than 200,000 images. Originally designed for face attribute recognition, each image is annotated with 40 attributes. Two commonly used subsets of CelebA in CRL research are CelebA(SMILE) and CelebA(BEARD). CelebA(SMILE) includes the attributes \textit{gender}, \textit{smile}, \textit{eyes open}, and \textit{mouth open}, while CelebA(BEARD) contains \textit{age}, \textit{gender}, \textit{beard}, and \textit{bald} attributes. The causal graphs corresponding to these subsets are illustrated in \cref{fig:CelebA}.

Both CelebA(SMILE) and CelebA(BEARD) exhibit several limitations that can misguide CRL models during training and hinder their generalization to real-world environments. First, the number of variables in both subsets is restricted—each dataset includes only four generative factors per image. In contrast, real-world data typically involve numerous interacting variables, both observed and unseen, which contribute to the complexity of causal structures. Consequently, datasets with a limited number of variables fail to expose CRL models to the level of causal diversity necessary for robust performance in realistic scenarios containing many known and unknown variables as well as confounders. Second, all variables in these datasets are binary. While categorical variables are indeed important for CRL development, restricting them to binary values represents an oversimplified form of categorical data. Furthermore, numerical variables, which are ubiquitous in real-world systems, are completely absent. This omission limits the model’s ability to learn and generalize to continuous domains, thereby reducing its applicability in practical tasks involving quantitative relationships. 

The most critical drawback, however, lies in the causal graphs associated with these datasets, which introduce several conceptual and methodological issues. Specifically, some of the assumed causal relationships lack empirical support or scientific validity. For example, in CelebA(BEARD), the variable \textit{age} is modeled as a cause of \textit{bald}, which is inaccurate since baldness is not deterministically dependent on age—many young individuals may be bald, while some older individuals may not be. Similarly, in CelebA(SMILE), \textit{gender} is considered a cause of \textit{eyes open}, an assumption that is neither biologically nor behaviorally justified and may even raise ethical concerns \cite{zhu2023shadow}. While correlations may exist among such variables, correlation does not imply causation, and these erroneous causal assumptions undermine the credibility of the dataset’s causal structure.  Although CelebA(SMILE) incorporates all three fundamental causal junction types—chain, fork, and collider—CelebA(BEARD) includes only fork (\eg $beard \gets age \to bald$) and collider (\eg $age \to bald \gets gender$) structures, while lacking any chain junctions. The absence of this essential structure reduces the dataset’s resemblance to real-world causal systems, where all three junction types commonly coexist.
In terms of confounders, CelebA(BEARD) again exhibits a fundamental structural limitation. A confounder, defined as a variable that simultaneously influences both the cause and the effect \cite{pearl2018book,vanderweele2013definition}, can be identified in CelebA(SMILE)—for instance, the variable \textit{smile} affects both the cause variable (\textit{mouth open}) and the effect variable (\textit{eyes open}). However, in the causal graph associated with CelebA(BEARD), no such confounding variable is present. This absence further diminishes the dataset’s alignment with real-world causal mechanisms, where confounders are typically unavoidable components of natural systems. These inaccuracies represent significant limitations of the CelebA datasets, which, despite their widespread use in CRL research, may lead to misleading causal representations and reduced model reliability in real-world applications.

\cite{zhu2023shadow} attempted to validate the causal relationships among variables in the CelebA(SMILE) and CelebA(BEARD) datasets by performing conditional independence tests, as recommended in \cite{pearl2009causality}. Specifically, they identified pairs of variables that should exhibit statistical dependence or conditional independence according to the junction types represented in the causal graphs of each dataset.
As discussed in \cref{sub: expected_features}, variables \textit{age} and \textit{gender} in CelebA(BEARD), and \textit{gender} and \textit{smile} in CelebA(SMILE), are expected to be statistically independent based on the causal structures provided. However, results from the $\chi^2$ independence tests indicated significant dependencies between these variable pairs, contradicting the assumed independence implied by the respective causal graphs. To address this inconsistency, \cite{zhu2023shadow} intentionally filtered the datasets to retain only samples in which these variables were statistically independent, thereby creating modified datasets that better aligned with the prescribed causal graphs.
However, we argue that the observed dependencies between variables expected to be independent could plausibly be attributed to latent confounders—unobserved variables simultaneously influencing both factors but omitted from the causal graphs (see \cref{fig:poss_conf}). This explanation suggests that the issue may not stem from incorrect causal assumptions but rather. However, this in turn raises broader concerns about the existence of hidden or disregarded confounders in these datasets—and potentially in other real-world datasets—which may compromise their causal integrity and, consequently, the reliability of CRL models trained on them.

In addition to the two commonly used versions of CelebA datasets, other versions are also constructed. For instance, in \cite{huang2025visual}, the authors created three distinct subsets of CelebA by selecting different combinations of attributes and establishing corresponding causal structures (see \cref{fig:celebA2}).
It is important to emphasize that the previously discussed limitations also exists in these versions. In particular, the causal relationships defined among the selected attributes should be mathematically justified or at least logically coherent, ensuring that the resulting dataset reflects plausible and scientifically meaningful causal mechanisms.

The aforementioned drawbacks have the potential to substantially mislead both causal discovery and causal inference in the developed models. As a consequence, the learned causal relationships may be incomplete, biased, or incorrect, which directly undermines the validity of downstream reasoning processes. In particular, counterfactual reasoning and counterfactual generation become unreliable, as they are grounded in flawed or poorly specified causal structures. This lack of reliability ultimately limits the practical applicability of such models, especially in real-world settings where robust and trustworthy causal explanations and interventions are essential.

\subsection{Synthetic Datasets}

Limitations mentioned for real-world datasets can be mitigated through the use of simulation-based synthetic data generation. Synthetic datasets—created under controlled conditions—offer desirable properties that make them well-suited for developing and evaluating CRL models. For instance, the number of variables can be systematically increased to create more complex and challenging learning scenarios, thereby improving the model’s robustness and generalizability. Furthermore, both numerical and categorical variables can be incorporated to ensure comprehensively covering all possible types of generative/cause-effect factors.
Synthetic data generation also allows for fine-grained variation in variable values, enabling the dataset to represent a broader range of conditions that may be infeasible to observe or manipulate in real-world settings (\eg altering a variable by one percent increments). Additionally, real-world data collection is often time-consuming, costly, and constrained by ethical or logistical factors, while synthetic data can be efficiently produced at scale. Also, a real world dataset will either require enforcing cause-effect relations or annotating which is very difficult.

Despite these advantages, synthetic datasets inherently lack the uncertainty and complexity of real-world environments. Consequently, CRL models trained solely on synthetic data may not fully generalize to practical applications. This challenge parallels the lab-to-deployment gap observed in other scientific and machine learning contexts—where models are first developed and validated under controlled (laboratory) conditions, and later adapted for real-world deployment through techniques such as domain adaptation or distributional shift adjustment. Similarly, CRL models can be initially trained on synthetic datasets and subsequently refined for real-world causal reasoning and inference.

\begin{figure*}
    \centering
    \includegraphics[width=0.65\linewidth]{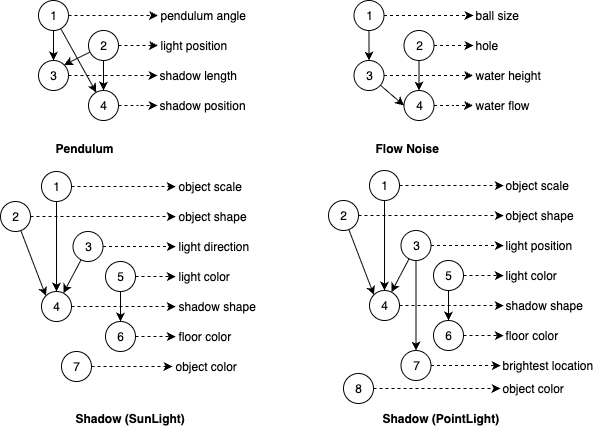}
    \caption{ausal graphs for Pendulum, Flow Noise, Shadow(SunLight), and Shadow(PointLight), four commonly used synthetic datasets for CRL model development.}
    \label{fig:synth_graphs}
\end{figure*}

Several synthetic datasets have been developed for CRL model evaluation and training, among which the most widely used are Pendulum \cite{yang2021causalvae}, Flow Noise \cite{yang2021causalvae}, Shadow(PointLight) \cite{zhu2023shadow}, and Shadow(SunLight) \cite{zhu2023shadow}. The causal graph corresponding to these datasets are illustrated in \cref{fig:synth_graphs}. Both Pendulum and Flow Noise datasets, originally introduced by \cite{yang2021causalvae}, consist of four numerical generative factors. While these datasets are valuable for foundational CRL research, they are limited by their small number of variables and lack of categorical factors, which restrict their ability to emulate the diversity and complexity of real-world data.
To overcome these limitations, \cite{zhu2023shadow} proposed two enhanced synthetic datasets—Shadow(PointLight) and Shadow(SunLight)—in which the number of generative variables is increased and categorical variables are explicitly incorporated, producing a more comprehensive and realistic setting for CRL model development.

Although the aforementioned synthetic datasets are inspired by real-world phenomena (\eg shadow formation and fluid dynamics), there remains a need for more sophisticated synthetic datasets that are more representative of real-world conditions and share a broader set of structural characteristics with them. In terms of basic causal junctions, the Pendulum dataset includes both fork (\eg \textit{shadow length} $\gets$ \textit{pendulum angle} $\to$ \textit{shadow position}) and collider (\eg \textit{pendulum angle} $\to$ \textit{shadow length} $\gets$ \textit{light position}) structures, but lacks a chain junction. The Flow Noise dataset, by contrast, contains chain (\eg \textit{ball size} $\to$ \textit{hole} $\to$ \textit{water flow}) and collider (\eg \textit{hole} $\to$ \textit{water flow} $\gets$ \textit{water height}) structures, but does not include a fork junction. Shadow(PointLight) includes fork and collider junctions but no chain, while Shadow(SunLight) contains only collider junctions and lacks both chain and fork structures.
With respect to the existence of confounders, none of these synthetic datasets include confounding variables when evaluated using the backdoor criterion \cite{pearl2018book}. Consequently, each of these datasets exhibits structural limitations that prevent them from being fully comprehensive.
While these datasets provide a valuable foundation for initial CRL model development, there remains a clear need for a comprehensive synthetic dataset that simultaneously incorporates all three fundamental causal junctions and confounding relationships. Such a dataset would offer a more realistic and robust benchmark for developing CRL models that generalize effectively to real-world scenarios.

\section{Evaluating Causal Inference Models}
\subsection{Dimensions to Evaluate}

 A CRL model should be capable of transforming the underlying generative factors of the input data into meaningful latent variables, disentangling these variables, identifying the causal dependencies among them, and leveraging this causal understanding to generate counterfactual samples. This comprehensive process encompasses several key components, each representing a distinct research direction. To ensure the model’s robustness and applicability in real-world scenarios, its performance should be systematically evaluated across all these dimensions. An effective CRL model is expected to demonstrate strong and balanced performance across all of these aspects.
 
In practice, however, many studies report model performance on only a subset of these evaluation dimensions, overlooking others. While certain CRL models may excel in one or two of these areas, a truly practical model must achieve consistent and high-quality results across all. For instance, a model that achieves accurate disentanglement of generative factors in the latent space but exhibits poor reconstruction capability would be unable to generate reliable counterfactual samples—rendering its disentanglement useless in downstream tasks. The same reasoning applies to other evaluation dimensions.

\begin{figure}
    \centering
    \includegraphics[width=0.7\linewidth]{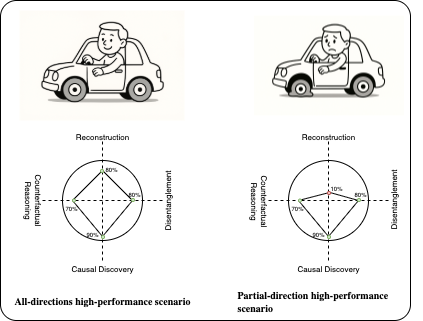}
    \caption{Illustration showing that if a CRL model performs well in all directions except one, its real-world applicability is compromised, analogous to a car with a single flat tire impairing overall system performance.}
    \label{fig:car_sample}
\end{figure}

Compared to a car, each evaluation direction functions like a tire—if one is deflated, the overall system fails to operate as intended (See \cref{fig:car_sample}) although the reset of the car might be functioning properly. Therefore, balanced performance across all axes is essential. For each of the aforementioned directions, the literature proposes a range of evaluation metrics, which have been employed by various studies. These metrics are discussed in detail in the following subsections.

\subsection{Reconstruction and Counterfactuals}
\label{sub: counter}
A CRL model should be capable of reconstructing the input data with high fidelity while also generating meaningful counterfactual samples through interventions in the latent space. Reconstruction quality is a fundamental prerequisite for reliable counterfactual reasoning. While strong reconstruction performance does not formally guarantee that the model can generate perfectly realistic samples, poor reconstruction ability typically indicates that the learned representation fails to accurately capture the underlying data distribution, which in turn limits the reliability of generated counterfactuals.

To quantitatively assess a model’s reconstruction capability, several metrics can be employed. Among these, mean absolute error ($MAE$) and mean squared error ($MSE$) are the most commonly used. Lower values of these metrics indicate better reconstruction performance. In cases where other evaluation metrics used in a study are oriented such that higher values correspond to better performance, it is recommended to normalize the reconstruction metrics to a common range (\eg $[0,1]$). Subsequently, one can transform them to align with this orientation by using $1-MAE$ or $1-MSE$, thereby ensuring consistency across all evaluation dimensions.

\begin{figure}
    \centering
    \includegraphics[width=0.7\linewidth]{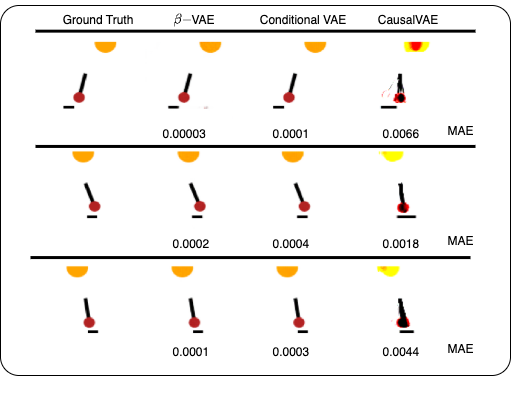}
    \caption{Reconstruction performance of different models. Although CausalVAE outperforms the other two models in MIC and TIC metrics, its reconstruction ability is inferior, highlighting a critical limitation for practical applications and underscoring the often-overlooked importance of reconstruction in CRL models}
    \label{fig:pend_gen}
\end{figure}

Despite its importance, reconstruction quality has often received comparatively less attention in CRL studies, which tend to focus more heavily on metrics evaluating disentanglement, causal discovery, or counterfactual reasoning. As a result, some models achieve strong performance on disentanglement-related metrics while exhibiting suboptimal reconstruction capability. For example, \cite{yang2021causalvae} report that the CausalVAE model outperforms baseline models such as $\beta$-VAE and conditional VAE on the Pendulum dataset in terms of MIC and TIC metrics for disentanglement. However, as illustrated in \cref{fig:pend_gen}, the reconstruction performance of CausalVAE is substantially lower than that of these baseline models, which may limit its ability to generate accurate and realistic counterfactual samples in practical applications.

Counterfactual samples can be generated by applying interventions to the latent variables, where in an ideal CRL model, each latent variable corresponds to a specific generative factor. Modifying a latent variable triggers updates to other variables according to the learned causal graph among the latent space variables. The evaluation of counterfactual reasoning can be performed along several criteria \cite{melistas2024benchmarking}, including composition, effectiveness, realism, and minimality. Composition assesses the model’s ability to reproduce the original sample, using metrics such as $L_1$ distance and Learned Perceptual Image Patch Similarity (LPIPS) \cite{zhang2018unreasonable}. Effectiveness measures whether the intervention achieved the intended effect. This can be quantified using anti-causal predictors trained on the input data distribution, and evaluated with metrics such as F1-score or mean absolute error (MAE) for classification and regression tasks. Realism evaluates whether the generated sample aligns with the input data distribution, using metrics such as Fréchet Inception Distance (FID) \cite{NIPS2017_8a1d6947}, Inception Score (IS) \cite{NIPS2016_8a3363ab}, or Kernel Inception Distance (KID). Minimality verifies that the counterfactual differs only with respect to the intervened variable, while all other variables remain unchanged. Metrics such as Counterfactual Latent Divergence (CLD) \cite{sanchez2022diffusioncausalmodelscounterfactual} are used for this purpose.

Although the aforementioned metrics can effectively evaluate a CRL model’s capability in counterfactual reasoning, two major challenges remain. First, most of these metrics are unbounded, making it difficult to normalize their values and report them alongside other directional evaluation metrics—such as those for reconstruction, disentanglement, and causal discovery. Second, these metrics do not directly assess the accuracy of the counterfactual samples generated by the model. In other words, while they ensure that the generated samples are close to the input data distribution, they do not guarantee that these samples are accurate representations of the expected outcomes.
Addressing this limitation is non-trivial, as evaluating the accuracy of generated counterfactuals requires access to ground truth labels, which are often unavailable or infeasible to obtain in many practical scenarios.

To address this challenge, several studies have proposed methodologies to evaluate the accuracy of generated counterfactual samples. For instance, in the work by \cite{10021114}, counterfactual samples were generated by applying interventions to the latent variables of their CRL model trained on synthetic datasets. Since the generative process of synthetic data is fully accessible, the same intervention can be applied directly to the data generation mechanism to produce the ground truth counterfactual sample. Having access to this ground truth enables the computation of reconstruction errors for the generated counterfactuals. While reconstruction-based metrics are informative, we recommend incorporating object detection–based metrics, such as the Jaccard index (Intersection over Union, IoU)(for this study the samples are images). This allows for a more detailed comparison between the generated counterfactual and the ground truth, assessing the spatial accuracy and structural correspondence of image components. This evaluation approach highlights a key advantage of using synthetic datasets, as they enable precise and quantitative assessment of counterfactual generation accuracy. Consequently, CRL models trained and validated on such datasets can be considered more reliable when deployed in real-world applications.

\subsection{Disentanglement}
A model’s disentanglement capability can be characterized along three principal properties. First, Modularity requires that each subset of latent space variables is exclusively associated with a specific subset of generative factors, with minimal influence on others. Second, Compactness emphasizes that the number of latent variables responsible for representing each generative factor should be as small as possible. Third, Explicitness demands that it should be possible to accurately recover the generative factors from the corresponding latent variables. To evaluate disentanglement, several supervised metrics have been proposed, including those based on mutual information \cite{NEURIPS2018_1ee3dfcd, do2021theoryevaluationmetricslearning,li2020progressivelearningdisentanglementhierarchical,sepliarskaia2021measuredisentanglement}, training simple predictive models \cite{8c5d4996ab134f8fa047c27b0fffb83c, NEURIPS2018_2b24d495, kumar2018variationalinferencedisentangledlatent}, and interventions on latent variables \cite{higgins2017beta, kim2018disentangling, kim2019relevance, suter2019robustly}. Additionally, information-theoretic metrics such as the Maximal Information Coefficient (MIC) and Total Information Coefficient (TIC) \cite{kinney2014equitability} are widely employed to assess CRL models’ disentanglement performance.

The disentanglement step is a critical component in CRL development, as it allocates distinct generative factors to specific latent variables. In a model with near-perfect disentanglement, editing a sample can be efficiently achieved by modifying the corresponding latent variables. Moreover, disentanglement has a direct impact on the causal discovery process. Therefore, the evaluation of disentanglement in a CRL model should be comprehensive, encompassing all baseline methods and addressing all three fundamental disentanglement properties—Modularity, Compactness, and Explicitness. In practice, rather than relying on a single metric, a suite of complementary evaluation metrics, each capturing a different aspect of the model’s disentanglement capability, should be employed.

We recommend always including MIC and TIC, as these metrics are widely used across studies in this field. Incorporating them facilitates direct comparison of results across different models and datasets. Additionally, we suggest using the Interventional Robustness Score (IRS) \cite{suter2019robustly}, an intervention-based metric that evaluates a CRL model’s performance in Modularity, Compactness, and Explicitness, which are essential for achieving effective disentanglement. As a predictor-based approach, DCI (Disentanglement, Completeness, and Informativeness) \cite{eastwood2018framework} provides a complementary assessment of the model along the same three disentanglement properties. Another information-theoretic metric, Joint Entropy Minus Mutual Information Gap (JEMMIG), offers a holistic evaluation of modularity, compactness, and explicitness. In summary, the set of metrics [MIC, TIC, IRS, DCI, and JEMMIG] is recommended for a comprehensive evaluation of a model’s disentanglement capability, encompassing multiple methodological perspectives and all key disentanglement criteria. All of these metrics are bounded between 0 and 1 and upward-oriented, meaning that higher values indicate better model performance.

\subsection{Causal Discovery}
After disentangling the latent variables, a crucial step in CRL models is to identify cause-effect relationships among them, enabling the model to perform causal inference. Causal discovery is an active research area \cite{pearl2018book}, and numerous techniques have been proposed over the years. Traditional causal discovery methods can be broadly categorized into constraint-based \cite{kalisch2007estimating, spirtes2013causal} and score-based \cite{chickering2002learning, yuan2013learning} approaches. In constraint-based methods, the skeleton of the underlying causal graph is first inferred using conditional independence tests, after which the edges are oriented according to a set of orientation rules. In score-based methods, each candidate graph is evaluated using a predefined scoring function, and the graph with the optimal score is selected. However, identifying the optimal graph is computationally intensive, and causal graphs obtained from both approaches are often not sufficiently accurate \cite{ng2019graph}. To address these limitations, Zhang et al. \cite{zheng2018dags} introduced a differentiable acyclicity constraint, reformulating causal discovery as a continuous optimization problem with a suitable loss function. While their method was primarily designed for linear relationships, subsequent studies \cite{yu2019dag, ng2019graph} extended it to effectively handle nonlinear relationships. More recently, the advent of large language models (LLMs) has enabled the development of LLM-based causal discovery techniques, offering new avenues for implementation \cite{10839116}.

The evaluation of a discovered causal graph is typically performed by comparing it with the ground truth causal graph derived from background knowledge. Commonly used metrics include Structural Hamming Distance (SHD), True Positive Rate (TPR), and Area Under the ROC Curve (AUC). SHD quantifies the difference between two graphs based on their adjacency matrices, accounting for three types of discrepancies: edges present in the ground truth but missing in the predicted graph, edges with incorrect orientation, and edges present in the predicted graph but absent in the ground truth. In contrast, TPR directly measures the proportion of true causal edges correctly identified by the model. Additionally, the Area Under the Curve (AUC) can be employed to assess the model’s overall ability to distinguish between true and false causal links.

It is important to note that causal graph discovery has a direct and substantial impact on the overall performance of a CRL model, as the model’s outputs are generated based on the extracted cause-effect relationships. Consequently, it is highly recommended to employ various techniques to quantify the uncertainty of the predicted causal graph, which in turn provides insight into the uncertainty of the model’s overall performance when generating outputs for practical applications.
Furthermore, some recent studies, such as \cite{huang2025visual}, bypass the full causal discovery step by leveraging background knowledge to provide the cause-effect relationships among latent variables. The model then focuses solely on learning the mapping function that best aligns with this predefined causal structure. This approach reduces uncertainty in the final outputs, making the results of the CRL model more reliable and robust.

\subsection{Metrics Unification}
While numerous CRL models have been proposed across different studies, their performance is typically reported in tabular form, where evaluation metrics across multiple directions are presented alongside baseline models for comparative analysis. However, these comparison tables can often be ambiguous and difficult to interpret, as models may outperform others in some dimensions while under-performing in others. Therefore, a dimensional comparison is essential—evaluating models separately across specific aspects such as reconstruction, disentanglement, causal discovery, and counterfactual reasoning—to determine which model performs best in each dimension. Even within a single evaluation direction, the use of multiple metrics can further complicate interpretation, since a model might excel on certain metrics but perform poorly on others within the same category. To provide a clearer understanding of this issue, we benchmarked three different models on the Pendulum dataset, employing multiple metrics for each evaluation dimension. The detailed results are presented in \cref{tabl:benchmarking}.

\begin{table*}[t] 
\caption{Performance of various models in different directions on pendulum dataset}
\resizebox{\textwidth}{!}{%
\begin{tabular}{lccccccccc}
\toprule

& & \multicolumn{4}{c}{Disentanglement} & \multicolumn{3}{c}{Counterfactual}\\
    \cmidrule(lr){3-6}
    \cmidrule(lr){7-9}
Models & Reconstruction↑ & IRS↑ & JEMMIG↑ & MIC↑ & TIC↑ & FID↓ & IS↑ & KID↓ \\
\midrule

$\beta$-VAE & 0.9863±0.0013 & \textbf{0.7965±0.0249} & 0.2556±0.0312 & 0.4133±0.1510 & 0.2802±0.1353 & 15.3481±3.3664 & 1.2137±0.0732 & 0.0136±0.0031\\
ConditionalVAE & \textbf{0.9881±0.0018} & 0.7559±0.0265 & 0.1930±0.0050 & 0.4331±0.0262 & 0.3174±0.0274 & \textbf{14.6477±3.9643} & 1.2083±0.0582 & \textbf{0.0117±0.0021} \\
 CausalVAE & 0.8654±0.0092 & 0.7087±0.0303 & \textbf{0.3538±0.0601} & \textbf{0.7409±0.0840} & \textbf{0.6038±0.0729} & 169.1046±14.0410 & \textbf{2.0548±0.2224} & 0.1995±0.0227 \\

\bottomrule
\end{tabular}
}
\label{tabl:benchmarking}
\end{table*}

As presented in \cref{tabl:benchmarking}, Conditional VAE demonstrates the best performance among the three models in the reconstruction direction (where reconstruction is computed as $1-MAE$). In terms of disentanglement, $\beta$-VAE achieves superior results under the IRS metric, while CausalVAE outperforms the other two models according to JEMMIG, MIC, and TIC scores. Regarding counterfactual generation, Conditional VAE again achieves the best performance based on FID and KID, whereas CausalVAE attains the highest IS value. Overall, these results indicate that it is not straightforward to determine which model consistently outperforms the others—either globally across all evaluation dimensions or even locally within a single dimension. This underscores the need for a unified or aggregate evaluation metric capable of facilitating comprehensive model comparisons across multiple evaluation dimensions.

\begin{figure}
    \centering
    \includegraphics[width=0.65\linewidth]{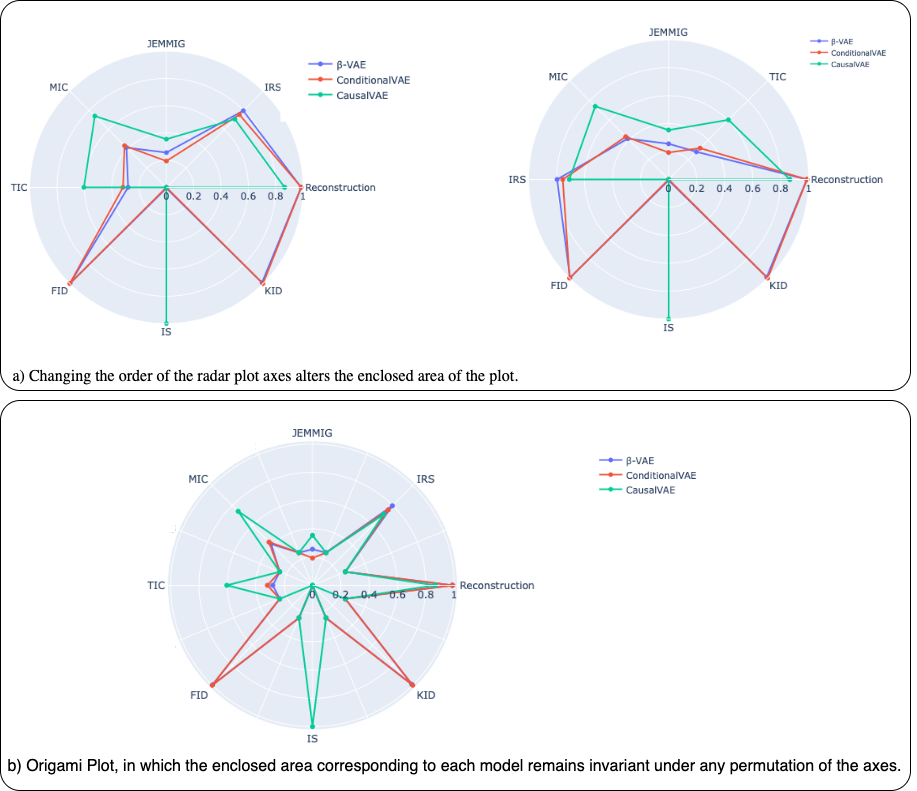}
    \caption{a) Radar plot for visual comparison of model performance across multiple directions and metrics. The shape and enclosed area are sensitive to the order of axes, making quantitative comparison potentially inconsistent. b) Origami plot for the same experiment, where the enclosed area for each model is insensitive to axis order, providing a more robust and reliable metric for overall performance comparison.}
    \label{fig:radar_origami}
\end{figure}

One practical approach for multi-dimensional model comparison is the use of a radar plot (see \cref{fig:radar_origami}). A radar plot provides an intuitive visualization of each model’s performance across multiple evaluation directions and metrics. However, for this visualization to be meaningful, all metrics must be normalized to the same range and oriented upward, meaning that higher values indicate better performance.

To enable radar plot generation for our experiments, we applied Min–Max normalization to the FID, IS, and KID metrics using their respective maximum and minimum values reported in \cref{tabl:benchmarking}. Since KID and FID are downward-oriented metrics, we transformed them using 1$-$FID and 1$-$KID to ensure consistency with the other metrics. The resulting radar plots allow for a visual and qualitative comparison of model performance across all evaluation directions. To obtain a single scalar value that consolidates and summarizes the information captured across all evaluation metrics, a promising approach is to compute the enclosed area of the radar plot corresponding to each model. For our experiments, the enclosed areas for $\beta$-VAE, Conditional VAE, and CausalVAE were 0.876, 0.855, and 0.556, respectively—suggesting that $\beta$-VAE, exhibits the best overall performance.
However, a key limitation of this approach is that the computed area is sensitive to the order of the axes in the radar plot. Altering the sequence of the metrics can change the geometric structure of the enclosed shape and thus the calculated area. For example, swapping the positions of IRS and TIC in \cref{tabl:benchmarking} results in a different radar plot (see \cref{fig:radar_origami}) and new enclosed areas of 0.907, 0.894, and 0.539 for $\beta$-VAE, Conditional VAE, and CausalVAE, respectively. In general, the enclosed area in radar plots is not invariant to axis order, making it an unreliable sole quantitative metric.

\begin{equation}
  \mathrm{S}_{radar}=\frac{Sin(\theta))}{2}\sum_{i=1}^{N}r_i.r_{i+1}, \qquad 
  \acute{\mathrm{S}}_{radar}=\frac{Sin(\theta))}{2}\sum_{i=1}^{N}\acute{r_i}.\acute{r_{i+1}},\qquad
  r_i\neq \acute{r_i}\longrightarrow \mathrm{S}_{radar}\neq\acute{\mathrm{S}}_{radar}
  \label{eq:S}
\end{equation}

Here, $\theta$ denotes the angle between two adjacent axes, while $r$ and $\acute{r}$ represent the metric values on each axis before and after permutation, respectively. Consequently, the single aggregated value reported by different studies for various models and datasets is inherently order-dependent, making the comparison across studies unreliable and non-standardized due to the sensitivity of the final result to the arrangement of metrics.

To mitigate this limitation, a modified visualization technique known as the Origami Plot \cite{duan2023origami} has been proposed, which ensures that the enclosed area remains invariant to the ordering of axes. In this approach, auxiliary axes are inserted between each pair of original axes, and a constant reference value is assigned to all auxiliary axes (see \cref{fig:radar_origami}). This modification effectively eliminates the dependency of the computed enclosed area on axis arrangement, thereby providing a more robust and order-insensitive comparative evaluation.

\begin{equation}
  S_{origami}=\sum_{i=1}^{N}2\frac{Sin(\frac{\theta}{2}).h.r_i}{2}=\sum_{i=1}^{N}Sin(\frac{\theta}{2}).h.r_i=Sin(\frac{\theta}{2}).h.\sum_{i=1}^{N}r_i
  \label{eq:orig_s}
\end{equation}

where $h$ denotes the reference value assigned to the auxiliary axes. The area enclosed by the Origami plot depends solely on the cumulative sum of the metric values and, therefore, remains invariant to the ordering of the metric axes. Consequently, we employ this area as a unified quantitative indicator that aggregates multiple evaluation metrics across different performance dimension of a CRL model. This approach provides an effective means of synthesizing diverse metrics into a single comprehensive score, facilitating overall model performance assessment and ranking. Although the Origami plot has been applied in other research domains \cite{Chen27072025, duan2023origami}, to the best of our knowledge, no prior work has utilized this or any comparable method to integrate CRL performance across multiple evaluation directions into a single quantitative index. In our experiments, the computed Origami scores (\ie the enclosed areas of the Origami plots) for $\beta$-VAE, Conditional VAE, and CausalVAE are 0.452, 0.448, and 0.409, respectively, with $h$=0.25. These scores can be further normalized by dividing them by the theoretical maximum enclosed area (0.765), corresponding to the case where all metric values equal 1, for this experiment including eight metrics. After normalization, the final Origami scores become 0.590, 0.586, and 0.534 for $\beta$-VAE, Conditional VAE, and CausalVAE, respectively.

\section{Discussion and Critics}

The development of CRL models has recently garnered significant attention, with numerous researchers focusing on advancing this line of work. Building such models requires careful consideration of several key factors, including the implementation of an effective causal discovery mechanism, the selection of an appropriate training dataset, and the execution of comprehensive evaluations across multiple directions to ensure the model’s robustness and applicability to real-world scenarios. This paper discusses these criteria in detail, highlights existing limitations within each aspect, and proposes potential solutions aimed at enhancing the generalizability and reliability of CRL models.

Moreover, several additional challenges must be addressed. Future CRL frameworks should be designed to perform effectively even with limited training samples, leveraging few-shot learning or other data-efficient techniques. As discussed by \cite{khemakhem2020variational}, unsupervised approaches for CRL often lead to identifiability issues, where two different sets of model parameters result in the same marginal distribution. To overcome this, CRL models should be developed in a supervised setting, where each sample is associated with a ground-truth label. However, this requirement introduces new difficulties, as the collection and annotation of large labeled datasets are both computationally expensive and time-consuming. Therefore, developing strategies that enable CRL models to learn effectively from a limited number of labeled samples is critical for addressing both the identifiability problem and the data scarcity challenge—particularly in medical domains, where access to extensive labeled datasets is inherently constrained \cite{schafer2024overcoming}.

Another major challenge in this domain is uncertainty quantification. Deep learning models are inherently susceptible to uncertainty, which must be explicitly measured and incorporated in studies related to CRL model development. This uncertainty can manifest at various stages of the CRL pipeline—from causal discovery to counterfactual reasoning. By performing uncertainty quantification, for instance through Bayesian neural networks \cite{pearl1995bayesian}, which aim to estimate probability distributions over model parameters rather than fixed point estimates, or by employing alternative probabilistic or ensemble-based uncertainty estimation methods \cite{abdar2021review,rahaman2021uncertainty}, the reliability and interpretability of CRL models can be significantly improved. Quantifying uncertainty enables practitioners to assess the confidence level of the model’s predictions and causal inferences, thereby enhancing the practical applicability and trustworthiness of CRL systems in real-world scenarios.

Furthermore, in the context of counterfactual reasoning, as discussed in \cref{sub: counter}, most existing evaluation metrics primarily assess whether the generated counterfactual samples belong to the same distribution as the training data. This ensures that the generated samples appear realistic and consistent with the data manifold learned during training. However, for real-world applications, it is essential to evaluate not only the realism but also the factual accuracy of the generated counterfactuals. A counterfactual image may indeed appear realistic and be statistically consistent with the training distribution, yet still fail to represent the true expected counterfactual outcome. In \cref{sub: counter}, we examined two commonly adopted methodologies for assessing the accuracy of counterfactual samples generated by CRL models, using both real-world and synthetic datasets.

\begin{figure}
    \centering
    \includegraphics[width=0.5\linewidth]{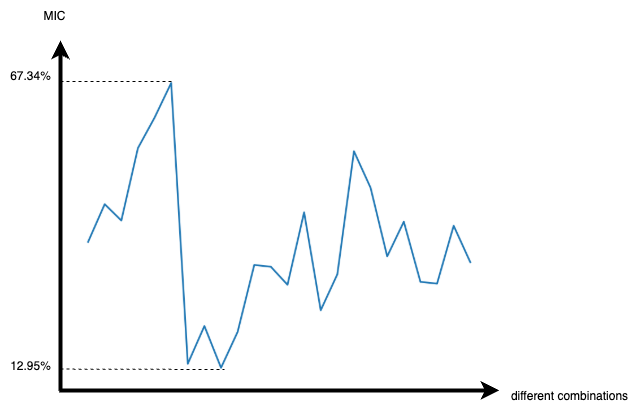}
    \caption{Different MIC values obtained by the $\beta$-VAE model trained on the Flow Noise dataset, where each result corresponds to a distinct assignment between the generative factors and the latent variables. This variation demonstrates the model’s high sensitivity to the permutation ambiguity inherent in unsupervised CRL frameworks.}    
    \label{fig: beta-vae}
\end{figure}

Another important consideration concerns the baseline models used in each study to benchmark the performance of newly developed CRL models. As discussed earlier, supervised learning is crucial in CRL due to the identifiability problem. Supervision provides explicit correspondence between the latent dimensions and the underlying generative factors in the input data—an association that cannot be established in an unsupervised framework. Consequently, models trained in an unsupervised manner often yield ambiguous latent representations, requiring additional interpretation to understand the learned factors. To illustrate this, consider a practical example using $\beta$-VAE \cite{higgins2017beta}, an unsupervised model, trained on the Flow Noise dataset for 100 epochs. Suppose the latent space consists of four dimensions, each ideally corresponding to one of four generative factors in the dataset. There exist 4! possible permutations of mappings between latent variables and generative factors, and each permutation yields a distinct evaluation outcome. \cref{fig: beta-vae} demonstrates how varying these assignments impacts the results. As shown, the MIC value can range from 12.95\%—when $\left[z_1, z_2, z_3, z_4  \right]$ correspond to [water height, ball size, water flow, and hole], respectively—to 67.34\%, when the order is [ball size, hole, water height, and water flow]. This example highlights the inherent ambiguity and sensitivity of evaluation results in unsupervised CRL models, emphasizing the importance of using supervised or partially supervised baselines for meaningful comparison.

Another major challenge in this field—one that has not been adequately or comprehensively addressed in existing studies—is the development of CRL models capable of handling categorical data. Categorical variables are an integral component of real-world data, and it is advisable to incorporate them into synthetic datasets as well. However, effectively modeling categorical data within CRL frameworks remains nontrivial. Most current studies treat categorical features as continuous numerical variables, which leads to conceptually and practically incorrect representations. For example, in the CelebA dataset, all attribute features are binary. Nevertheless, many CRL-based studies treat these binary features as numerical and, during counterfactual generation, vary their values over a continuous range. This approach is fundamentally flawed because binary variables can only assume two discrete values. To address this issue, it is recommended to employ one-hot encoding and integrate suitable experimental strategies that ensure stable and efficient forward and backward propagation during training. A CRL model designed with such considerations can more faithfully capture the discrete nature of categorical data, resulting in more reliable, interpretable, and semantically valid counterfactual representations.

\subsection{Reproducibility}
Similar to other deep learning architectures, CRL models are implemented using various computational frameworks. Ensuring the reproducibility of these models requires the public release of source code and comprehensive documentation of experimental settings. This practice offers several key advantages. First, it enables other researchers to replicate the reported results and independently validate the findings and claims of the original study. Second, it facilitates research continuity and model improvement, allowing new methods to be developed upon existing implementations—for example, SCM-VAE \cite{10021114} was developed using CausalVAE \cite{yang2021causalvae} as its foundational framework. However, in the current landscape of CRL research, there exists a significant reproducibility gap. Many studies fail to fully release their implementation details, and the available codebases are often incomplete, insufficiently documented, or inconsistent with the descriptions provided in the corresponding papers. As a result, even when researchers attempt to reimplement published models, the outcomes deviate from the results originally reported, thereby hindering progress and transparency in the field.

To illustrate this issue, consider the CausalVAE study, whose implementation has been made publicly available on GitHub\footnote{https://github.com/huawei-noah/trustworthyAI/tree/master/research/CausalVAE}. In this work, the authors developed and evaluated their model on four datasets: two synthetic datasets (Pendulum and Flow Noise) and two real-world datasets (CelebA(SMILE) and CelebA(BEARD)). However, the released codebase only includes the model architectures and implementation details corresponding to the two synthetic datasets. The absence of code and configuration files for the CelebA causal datasets makes it impossible to reproduce or validate the model’s reported results on real-world data. Furthermore, even for the portions of the code that were released, the reported performance metrics are not consistently reproducible. For example, the original study reports MIC and TIC values of 95.1±2.4 and 81.6±1.9 on the Pendulum dataset, and 72.1±1.3 and 56.4±1.6 on the Flow Noise dataset. To ensure a fair comparison, we reimplemented the publicly available code from the CausalVAE repository and conducted experiments across 10 independent runs to mitigate random variance. The aggregated results, alongside those reported in the original CausalVAE paper, are summarized in \cref{tab:repreducibility}. Specifically, the table reports: (i) results averaged over all 10 runs (All), (ii) results obtained after excluding the highest and lowest values (Boundaries Out), and (iii) performance corresponding to the top 5 runs based on metric values. Note that all the values are calculated using the Hungarian-algorithm-based \cite{kuhn1955hungarian,kuhn1956variants} matching scheme.

This critique extends to several other studies in the field. Some have not released their implementation code to date, while others have published incomplete or partially impaired versions, as previously discussed. Another major reproducibility challenge arises from the inconsistent use of the CelebA causal datasets across different studies. Specifically, various works have employed different subsets of the dataset for training, leading to inconsistencies in experimental settings and reported results. For instance, CausalVAE \cite{yang2021causalvae} utilized 20K samples from the dataset, SCM-VAE \cite{10021114} also used 20K samples but selected them randomly (and not necessarily the same subset), CFI-VAE \cite{li2024causal} trained on 30K randomly selected samples, and CIDiffuser \cite{huang2025visual} employed nearly the entire dataset consisting of over 200K samples. Such discrepancies in the number and selection of training samples—combined with the lack of publicly shared dataset splits—significantly reduce the reproducibility of the reported findings. Moreover, these inconsistencies make it difficult to conduct fair and standardized comparisons across models, as variations in dataset composition can heavily influence both the training dynamics and the final performance metrics.

\begin{table}[]
    \centering
    \caption{MIC and TIC values of CausalVAE reported by the original study (column CausalVAE), and the results after our reimplementation.}
    \resizebox{\textwidth}{!}{%
    \begin{tabular}{ccccccccc}
    \toprule
    & \multicolumn{4}{c}{Pendulum} & \multicolumn{4}{c}{Flow Noise}\\
    \cmidrule(lr){2-5}
    \cmidrule(lr){6-9}
    &CausalVAE&All&Boundries out&Top 5& CausalVAE&All&Boundries out&Top 5\\
    \midrule
         MIC&95.1±2.4 &69.1± 21.7 &74.7±6.9 &81.2±5.9 &72.1±1.3 &57.3±26.6 &61.1±22.7 & 74.1±1.0\\
         
         TIC& 81.6±1.9 &56.0±19.5 &60.7±6.6 &66.9±6.7 &56.4±1.6 & 47.9±25.4& 51.4±21.9 & 64.6±1.2\\
    \bottomrule
    \end{tabular}
    }
    
    \label{tab:repreducibility}
\end{table}

\bibliography{main}
\bibliographystyle{other/tmlr}


\end{document}